# Text Analysis Using Deep Neural Networks in Digital Humanities and Information Science

Omri Suissa[1,2] | Avshalom Elmalech[1] | Maayan Zhitomirsky-Geffet[1]

[1] Bar Ilan University, Department of Information Science, Ramat Gan 52900, Israel

[2] Correspondence email: omrishsu@gmail.com

## Abstract

Combining computational technologies and humanities is an ongoing effort aimed at making resources such as texts, images, audio, video, and other artifacts digitally available, searchable, and analyzable. In recent years, deep neural networks (DNN) dominate the field of automatic text analysis and natural language processing (NLP), in some cases presenting a super-human performance. DNNs are the state-of-the-art machine learning algorithms solving many NLP tasks that are relevant for Digital Humanities (DH) research, such as spell checking, language detection, entity extraction, author detection, question answering, and other tasks. These supervised algorithms learn patterns from a large number of "right" and "wrong" examples and apply them to new examples. However, using DNNs for analyzing the text resources in DH research presents two main challenges: (un)availability of training data and a need for domain adaptation. This paper explores these challenges by analyzing multiple use-cases of DH studies in recent literature and their possible solutions and lays out a practical decision model for DH experts for when and how to choose the appropriate deep learning approaches for their research. Moreover, in this paper, we aim to raise awareness of the benefits of utilizing deep learning models in the DH community.

## Introduction

The research space of digital humanities (DH) applies various methods of computational data analysis to conduct multi-disciplinary research in archaeology (Eiteljorg, 2004; Forte, 2015), history (Thomas, 2004; Zaagsma, 2013), lexicography (Wooldridge, 2004), linguistics (Hajic, 2004), literary studies (Rommel, 2004), performing arts (Saltz, 2004), philosophy (Ess, 2004), music (Burgoyne, Fujinaga, & Downie 2015; Wang, Luo, Wang, & Xing, 2016), religion (Hutchings, 2015) and other fields. The scope of DH continues to expand with the development of new information technologies, and its boundaries remain amorphous (McCarty, 2013). Therefore, DH's definition is unclear and may have different interpretations (Ramsay, 2016; Poole, 2017). Library and Information Science (LIS) and DH research have a similar and overlapping scope and interfaces (Posner, 2013; Koltay 2016), to the extent that some propose to integrate and combine both research fields (Sula, 2013; Robinson, Priego, & Bawden, 2015). DH and LIS academic units are often located together (Sula, 2013), and share a significant volume of common topics, such as metadata, linked data and ontologies, information retrieval, collection classification, management, archiving and curation, bibliographic catalogue research, digitization of printed or physical artifacts, preservation of cultural heritage, data mining and visualization, and bibliometrics (Svensson, 2010; Russell 2011; Gold, 2012; Warwick 2012; Sula, 2012; Beaudoin, & Buchanan, 2012; Sula 2013; Drucker, Kim, Salehian, Bushong, 2014; Koltay, 2016; Gold, & Klein 2016). However, regardless of the definition or research scope, many (if not most) of the research in DH/LIS focuses on textual resources, recorded information, and documents (Robinson et al., 2015; Poole, 2017). Therefore, this paper argues that a deep understanding of text analysis methods is a fundamental skill that future (and present) DH/LIS experts must acquire.

Supervised deep neural networks (deep learning) are a subset of machine learning algorithms considered to be the state-of-the-art approach for many NLP tasks, such as entity recognition (Li, Sun, Han, & Li, 2020), machine translation (Yang, Wang, & Chu, 2020), part-of-speech tagging and other tasks (Collobert & Weston, 2008) from which many DH/LIS text analysis research projects can benefit. Therefore, this paper aims to raise the awareness of DH and LIS researchers of state-of-the-art text analysis (NLP using deep neural networks) approaches and techniques. This is not the first attempt to make NLP technologies accessible or highlight the benefits of NLP to the DH/LIS research community (Biemann, Crane, Fellbaum, & Mehler, 2014; Kuhn, 2019; Hinrichs, Hinrichs, Kübler, & Trippel, 2019; McGillivray, Poibeau, & Ruiz Fabo, 2020). However, this paper argues that in addition to bridging between the NLP community and the DH/LIS research community, the DH/LIS research community should cultivate experts with a deep understanding of the technological space, experts that are capable of customizing and developing the technology themselves. Use of "off the shelf" tools and algorithms is no longer sustainable (Kuhn, 2019); the future DH expert must be comfortable using and adapting state-of-the-art NLP methodologies and technologies to the DH-specific tasks. To the best of our knowledge, this is the first attempt to highlight the challenges and analyze the potential solutions of the common usage of deep neural networks for text analysis in the DH/LIS space.

DNN models are often developed by computer scientists and trained, tested, and optimized for generic, open-domain tasks or by commercial enterprises for modern texts (Krapivin, Autaeu, & Marchese, 2009; Rajpurkar, Zhang, Lopyrev, & Liang, 2016). However, applying these DNN models for DH/LIS tasks and textual resources is not straightforward and requires further investigation. This paper presents the practical challenges that DH/LIS experts may encounter when applying DNN models in their research by examining multiple use cases presented in current literature, alongside an overview of the possible solutions, including deep learning technology. Although there might be other methodological challenges (Kuhn, 2019), this paper focuses on the two main practical challenges faced when applying deep learning for almost every DH research:

(1) Training data (un)availability - DH text resources are often domain-specific and niche, and contain a relatively small number of training examples; thus, there is not enough data for the DNN learning process to converge. Even when there is a large DH text corpus, there are no balanced ground truth labeled datasets (i.e., datasets with the distribution of "right" and "wrong" examples representative of the corpus) from which the DNN can learn (McGillivray et al., 2020), and changes or adaptations in the network architecture are required in order to achieve high accuracy for such datasets (Hellrich & Hahn, 2016).

(2) Domain adaptation - in many tasks considered "common" in NLP, the DH interpretation of the task is different from the standard interpretation. Moreover, DH text resources may need to be preprocessed before serving as input to DNNs, due to "noisy" data (biased, contains errors or missing labels or data (Hall, 2020; Prebor et al., 2018)) or non-standard data structure, such as mixed data formats (combining unstructured text, semi-structured and structured data in the same resource). In many cases, these resources are unsuitable for serving as an input into DNN models, or if they are used as-is, the models do not achieve maximum accuracy.

These challenges have unique implications on the utilization of DNNs with DH/LIS resources and tasks and, in various cases, may require different solutions. As a result of this study, a decision model for choosing the appropriate machine-learning approach for DH/LIS research is presented as a practical guideline for experts, with topics that digital humanitists should master being outlined.

## Digital Humanities and Automatic Text Analysis

Natural Language Processing (NLP) is a research area that explores how computational techniques (algorithms) can be used to understand and transform natural language text into structured data and knowledge (Young, Hazarika, Poria, & Cambria, 2018; Chowdhary, 2020). Until a few years ago, the state-of-the-art techniques that addressed supervised natural language processing challenges were based on a mix of machine learning algorithms. NLP tasks such as text classifications, entity recognition, machine translation, and part-of-speech tagging were solved using various classic supervised machine learning algorithms, such as Support Vector Machine (SVM), Hidden Markov Model (HMM), decision trees, k-nearest neighbors (KNN), and Naive Bayes (Zhou & Su, 2002; Liu, Lv, Liu, & Shi, 2010; Vijayan, Bindu, & Parameswaran, 2017). Basically, these algorithms apply a manually selected set of characteristic features to a given task and corpus, and a labeled dataset with "right" and "wrong" examples for training the optimal classifier. Given a new example of the same type, this classifier will be able to automatically predict whether or not this example belongs to the predefined category (e.g., whether a given sentence has a positive sentiment or not).

However, in many cases, it is not easy to decide what features should be used. For example, if a researcher wishes to learn to classify a text's author from the Middle Ages, she will need to use the features that represent the unique writing styles that distinguish the authors. Unfortunately, it is not easy to describe these features in terms of textual elements. Deep learning solves this central problem by automatically learning representations of features based on examples instead of using explicit predefined features (Deng & Liu, 2018). Deep learning (DL) is a sub-field of machine learning that draws its roots from the Neurocognition field (Bengio, Goodfellow, & Courville, 2017). The DL approach uses deep neural networks (DNN) models for solving a variety of Artificial Intelligence tasks. The technical details of various DNN models and techniques appear in Appendix I.

DH researchers use NLP algorithms for DH-specific tasks in various domains. For example, Niculae, Zampieri, Dinu, and Ciobanu (2014) used NLP techniques to automatically date a text corpus. They developed a classifier for ranking temporal texts and dating of texts using a machine learning approach based on logistic regression on three historical corpora: the corpus of Late Modern English texts (de Smet, 2005), a Portuguese historical corpus (Zampieri & Becker, 2013) and a Romanian historical corpus (Ciobanu, Dinu, Dinu, Niculae, & Sulea, 2013). To construct social networks among literary characters and historical figures, Elson, Dames, and McKeown (2010) applied "off-the-shelf" machine learning tools for natural language processing and text-based rules on 60 nineteenth-century British novels. Zhitomirsky-Geffet and Prebor (2019) used lexical patterns for Jewish sages disambiguation in the Mishna, and then applied several machine learning methods based on Habernal and Gurevych's (2017) approach for the co-occurrence of sages and pattern-based rules for specific inter-relationship identification in order to formulate a Jewish sages social interactions network. In paleography, the study of historical writing systems and the deciphering and dating of historical manuscripts, Cilia, De Stefano, Fontanella, Marrocco, Molinara, and Freca (2020) utilized MS-COCO (Lin, Maire, Belongie, Hays, Perona, Ramanan, & Zitnick, 2014), a generic corpus of images, and a domain-specific corpus to train DNN models and design a pipeline for medieval writer identification. To predict migration and location of manuscripts, Prebor, Zhitomirsky-Geffet and Miller (2020a, 2020b) devised lexical patterns for disambiguation of named entities (dates and places) in the corpus of the Department of Manuscripts and the Institute of Microfilmed Hebrew Manuscripts in the National Library of Israel. Next, the authors trained a CART machine learning classifier (Classification and regression tree based on Decision Tree learning) (Rokach and Maimon, 2015) to predict the places of manuscripts that were often absent in the corpus. For ancient languages analysis, a study (Dereza, 2018) compared accuracy for lemmatization for early Irish data using a rule-based approach and DNN models, and proved the advantages of using DNN on such a historical language - even with limited data. For historical network analysis, Finegold, Otis, Shalizi, Shore, Wang, and Warren (2016) used named entity recognition tools (Finkel, Grenager, & Manning, 2005; Alias-i, 2008) with manual rules on the Oxford Dictionary of National Biography and then applied a regression method, namely Poisson Graphical Lasso (Yang, Ravikumar, Allen & Liu, 2013) to find correlations between entities (nodes). Nevertheless, as demonstrated by the examples above, although there is a "computational turn" (Berry, 2011) in the DH research and methodologies, state-of-the-art computational NLP algorithms, like deep neural networks, are still rarely used within the core research area of DH (Kuhn, 2019).

To estimate the potential of deep learning use in DH, a comparison has been performed to one of the fields that is similar to DH - Bioinformatics. These fields are comparable since both are characterized by their inter-disciplinarity and because Bioinformatics thrives on application of computational analysis for exploring and investigating information repositories in a chosen knowledge domain (Ewens & Grant, 2006). A list of leading journals was compiled in each field and searched for articles with "deep neural network" and "machine learning" keywords. For DH, twelve journals were selected, based on Spinaci, Gianmarco, Colavizza, Giovanni, & Peroni (2019), all in English and ranked as 1 (exclusively DH). For Bioinformatics, twelve journals were selected based on Google Scholar's top publication list[1]. The two lists of the journals appear in Appendix III.

The comparison was conducted on the articles published in the above journals over the past three years and measured the following: 1) the percentage of articles with each of the two keywords in the selected journals in each field, to ascertain the usage of machine learning (ML) in general vs. deep learning (DL) in particular, in each field; and 2) the percentage of articles mentioning deep learning out of the machine learning articles in each field. As can be observed from Figure 1, in the DH field, only 21% of the articles discussing "machine learning" also discussed "deep learning"; while in Bioinformatics, 52% of the articles discussing "machine learning" also discussed "deep learning". Moreover, in the DH field, only 3.8% of the articles mentioned "deep learning", while in Bioinformatics, 19.5% of the articles mentioned "deep learning" – five times higher. In addition, in the DH field, 18% of the articles discussed "machine learning", while in Bioinformatics, 37% of the articles discussed "machine learning" – only two times higher. These results indicate that the DH field "lags behind" when it comes to using machine learning and especially deep learning state-of-the-art models.

---

[1] https://scholar.google.com/citations?view_op=top_venues&hl=en&vq=bio_bioinformatics

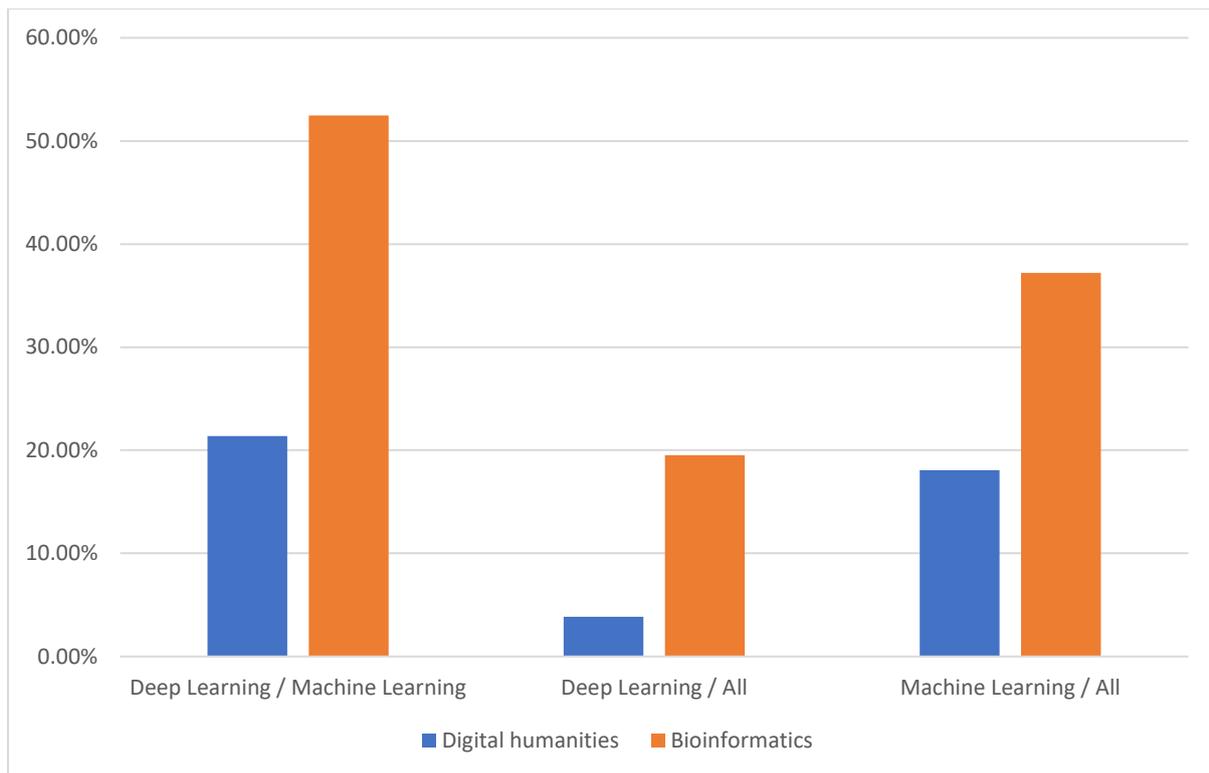

*Figure 1: Deep neural networks and machine learning articles in DH/LIS vs. Bioinformatics.*

The next section provides an in-depth analysis of challenges and potential solutions for using DNN in DH/LIS, supported by multiple use-case studies from the recent DH literature. The analysis is divided into two main sections dealing with two primary challenges in applying deep learning to DH research: training data (un)availability and domain adaptation.

## Challenges when Using Deep Learning for Digital Humanities Research

### Training Data (Un)availability

Computer scientists often work on generic supervised text analysis tasks with open-domain or modern datasets. Kaggle[2], the machine learning community, hosts many of these datasets. For example, the IMDb dataset contains a short description of a movie, and its review score allows to research sentiment analysis (Maas, Daly, Pham, Huang, Ng, & Potts, 2011); question answering system can be developed using Stanford Question Answering Dataset (Rajpurkar, Jia, & Liang, 2018); and SPAM filtering can be developed using a dedicated dataset (Almeida, Hidalgo, & Silva, 2013). Unfortunately, the DH community has not (as yet) produced large annotated open datasets for researches (although there are few in niche areas like (Rubinstein, 2019; Chen & Chang, 2019)). The lack of annotated data is a challenge for both classical machine learning and deep learning supervised algorithms (Elmalech & Dishi 2021). However, supervised deep learning algorithms require significantly more data than machine learning algorithms, making this challenge a critical practical challenge for DH researchers. This is one reason that even when DH/LIS researchers use deep learning, they often use unsupervised algorithms that do not require training data and are limited to specific tasks (Moreno-Ortiz, 2017). This section investigates some of the methods that DH researchers can apply to overcome this challenge.

---

[2] https://www.kaggle.com/datasets

**Training Dataset Generation by Humans**

Humans are the best alternative for dataset generations due to their domain knowledge and high accuracy. Therefore, the first consideration when generating a dataset is to consider if humans can be used for the job. However, humans are not as scaleable as computer software. It is possible to manually generate a dataset by humans when the needed labeling is relatively small or as a baseline for synthetic dataset generation. There are two types of manual dataset generation: crowd-based dataset generation and domain expert-based dataset generation. Crowdsourcing dataset generation is a relatively cheaper and effective method, but it can only be used when the labeling is "common knowledge". In some cases, for example, in the study aiming to generate a dataset of relationships extraction between characters in literary novels (Chaturvedi et al., 2016), the researchers must use expert annotators that can read and understand a novel; or even annotate themselves when working with historical languages known only to a few, as in Schulz & Ketschik (2019).

Crowdsourcing is based on large groups of non-expert, low-paid workers or volunteers performing various well-defined tasks. Existing studies tested optimization strategies for different tasks, such as extracting keyphrases (Yang, Bansal, Dakka, Ipeirotis, Koudas, & Papadias, 2009), natural language and image annotation (Snow, O'Connor, Jurafsky, & Ng, 2008; Sorokin & Forsyth, 2008), and document summarization (Aker, El-Haj, Albakour, & Kruschwitz, 2012). Crowdsourcing requires quality control to ensure that crowd workers are performing their tasks at a satisfactory level (Elmalech Grosz 2017). One of the effective generic (task-agnostic) quality control techniques is filtering out tasks with a low inter-worker agreement (Bernstein, Little, Miller, Hartmann, Ackerman, Karger, Crowell, & Panovich, 2010; Downs, Holbrook, Sheng, & Cranor, 2010; Kittur, Smus, Khamkar, & Kraut, 2011). Another popular approach is breaking tasks into sub-tasks (Bernstein et al., 2010; Kittur et al., 2011).

Employing crowd workers for dataset generation has been carried out in various domains, including DH projects (e.g., Elson, Dames, & McKeown, 2010). Thus, in this use-case study, Elson et al. (2010) utilized crowdsourcing to build a dataset of quoted speech attributions in historical books in order to generate a social network among literary characters. Elson et al. (2010) did not use DNN, but rather classic machine learning methods (Davis, Elson, & Klavans, 2003), but the dataset generating process is the same for classic ML and DL.

Another example of such a use-case is fixing Optical Character Recognition (OCR) errors in historical texts. In the DH/LIS space, there is great interest in investigating historical archives. Therefore, over the past few decades, archives of paper-based historical documents have undergone digitization using OCR technology. OCR algorithms convert scanned images of printed textual content into machine-readable text. The quality of the OCRed text is a critical component for the preservation of historical and cultural heritage. Unsatisfactory OCR quality means that the text will not be searchable, analyzable, or analysis may result in wrong conclusions. Unfortunately, while generic OCR techniques and tools achieve good results on modern texts, they are not accurate enough when applied to historical texts. Post-correction of digitized small scale or niche language historical archive is a challenge that can be solved using DNNs with high accuracy (Chiron, Doucet, Coustaty, & Moreux, 2017; Rigaud, Doucet, Coustaty, & Moreux, 2019) if an appropriate dataset is attainable. Therefore, the first thing that should be researched is an effective methodology for crowdsourcing this specific task (Suissa, Elmalech, & Zhitomirsky-Geffet, 2019). The details of the crowdsourcing research are outside the scope of this paper. What is essential from the DH/LIS research point of view is that the findings of Suissa et al. (2019) proved to be an effective dataset generation approach. Using the developed strategies, DH researchers can optimize the process to achieve better results matching their objectives and priorities. The corrected corpus of OCRed texts created by the optimized crowdsourcing procedure can serve as a training dataset for DNN algorithms.

However, although the crowdsourcing method yields satisfactory results, it is suitable mainly for widely spread languages like English or Spanish. Other national languages do not have enough crowd workers-speakers to utilize such an approach effectively. Moreover, manually generating a dataset for training a DNN model in order to post-correct OCR errors is expensive and inefficient, even when the task is crowdsourced. Therefore, in practice, this human-only dataset generation should be shifted to a human-in-the-loop solution.

**Training Dataset Generation using Algorithms**

The next range of solutions takes a two-phase approach. In the first phase, humans are used to create a small set of examples; this set of examples is used in the second phase by a different set of algorithms to generate a synthetic dataset with numerous training examples (Pantel, & Pennacchiotti, 2006; Bunescu, & Mooney, 2007). One way is to find recurring patterns in a small number of manually corrected examples, and use them to generate more correct examples. Thus, the use-case study that adopted this approach for automatic training dataset generation in the OCR post-correction domain, Suissa, Elmalech, & Zhitomirsky-Geffet (2020) used crowd workers to fix a relatively small set of OCRed documents. Then, the Needleman–Wunsch alignment algorithm (Needleman, & Wunsch, 1970) was used to find common confusions between characters committed by the crowd workers.

Using this confusion list, a large dataset of "wrong" and "right" sentences was generated and used by a DNN to correct historical OCRed text.

Another way to generate a dataset from a small set of manual examples is called "distant supervision" (Mintz, Bills, Snow, & Jurafsky, 2009). In this approach, a classifier is trained on a small set of examples and is applied to a large corpus. The classifier will classify the data with a relatively low accuracy but sufficiently high accuracy for the DNN to learn other features from this weak classification. Blanke, Bryant, & Hedges (2020) used this method to perform sentiment analysis on Holocaust testimonials data (Thompson, 2017). In the first phase, they did not use crowd workers for the initial dataset generation but rather applied a dictionary-based approach to find negative and positive sentiment sentences based on the TF-IDF measure (Singhal, 2001). Using these sentences, they trained a classifier to distinguish between positive and negative examples. In the second phase, they used the classifier to produce a large training corpus of positive and negative memories of Holocaust survivors for DNN text analysis. Using this method eliminates the need for humans; however, it is suitable only for specific tasks.

A different approach to solving the training dataset's unavailability is the transfer learning (Torrey, & Shavlik, 2010) method. In transfer learning, a generic dataset is used; the dataset should be suitable for the task needed to be solved, but with open-domain / other domain data. The model is then trained again using a small set of domain-specific examples (generated by humans or artificially). This approach is based on the intuition that humans transfer their knowledge between tasks based on previous experiences. Cilia et al. (2020) utilized transfer learning to identify medieval writers from scanned images. Instead of generating a large dataset, they used a model that was already trained on an open generic dataset MS-COCO (Lin et al., 2014) and trained it again using a small set of domain-specific examples from the Avila Bible (images of a giant Latin copy of the Bible). Banar, Lasaracina, Daelemans, & Kestemont (2020) applied transfer learning to train neural machine translation between French and Dutch on digital heritage collections. They trained several DNNs on Eubookshop (Skadiņš, Tiedemann, Rozis, & Deksne, 2014), a French-Dutch aligned corpus. Then, instead of training the DNN models directly on the target domain data, they first trained the models on "intermediate" data from Wikipedia (articles close to the target domain). Only then did they train the models for the third time on the target domain data - the Royal Museums of Fine Arts of Belgium dataset. Using this "intermediate fine-tuning" approach, Banar et al. (2020) achieved high accuracy for French-Dutch translation in the domain of Fine Arts. This method can also solve another challenge for the DH/LIS researcher when using DNN models – the domain adaptation challenge.

Recent studies (Radford, Wu, Child, Luan, Amodei, & Sutskever, 2019; Brown, Mann, Ryder, Subbiah, Kaplan, Dhariwal, & Amodei, 2020) show that in some cases, instead of fine-tuning a pre-trained model, a large-scale pre-trained model, such as GPT3 (Radford et al., 2019), trained on ~500 billion (modern) words, can achieve good results with a limited (or without) domain-specific dataset. Although these methods (named Few-shot and Zero-shot learning) do not reach the same performance as the fine-tuning method, they are preferable for low resource domains when dataset generation is impossible. However, most of the models that are pre-trained on a large-scale modern English dataset and suitable for Few-shot and Zero-shot learning may not reach the same accuracy for DH historical corpora, especially in (other than English) national languages, due to a bias towards modern language.

**Domain Adaptation**

Even with a large dataset ready for DNN training, there are other challenges a DH/LIS expert may encounter when attempting to solve a text analysis task on DH/LIS data with DNNs. As mentioned in the previous section, data is a critical part of DNN's high accuracy. However, specific task/domain adaptation is just as vital, and without adapting the model or the architecture to the specific task and domain, the DNN may perform poorly.

A DNN model is a set of chained mathematical formulas with weights assigned to each node (neuron) expressing a solution to a specific task. Although there are regulation techniques to generalize the DNN model, in many cases training the model with different data will significantly impact the weights. In other words, using the same mathematical formulas, the learning process interprets the same task differently. In this context, transfer learning described in the previous section can also serve as a domain adaptation method, since the DNN model's interpretation of the task is adjusted to the domain-specific data. Moreover, DH/LIS text analysis tasks are not just different in terms of interpretation but also often require domain-specific preprocessing and analysis pipeline. Therefore, in order to improve the accuracy of DNN models for text analysis tasks, DH/LIS experts should be familiar with methods and techniques for customizing DNN models, preprocessing DH/LIS data, and adapting the analysis pipeline.

**DNN Optimization for DH-specific Tasks**

A DNN model has a high number of architecture components and hyper-parameters that influence the model training efficacy and accuracy. Selecting the domain-specific suitable components and hyper-parameter values may considerably improve the performance of the DNN (Bengio, 2012). Here are a few of the most common architectures and hyper-parameters that an expert should consider (see Appendix I for technical details):

- Architecture components:
    - Type of the model – for instance, RNN-based, SAN-based (Vaswani et al., 2017), feed-forward-based, Transformers-based (Devlin et al., 2018).
    - Type and size of the layers – including individual layers, such as CNN (Albawi, Mohammed, & Al-Zawi, 2017), LSTM (Hochreiter et al., 1997), GRU (Cho et al.,

2014), ResNet (He, Zhang, Ren, & Sun, 2016), AlexNet (Krizhevsky, Sutskever, & Hinton, 2012), and multi-layer architectures, such as BERT (Devlin et al., 2018). These can be applied with or without bidirectionality (Schuster et al., 1997), attention (Bahdanau, Cho, & Bengio, 2015), skip-connections (Chang, Zhang, Han, Yu, Guo, Tan, & Huang, 2017), and other architectural components.

- o Type of input - DNN input is a vector (a series of numbers). Each number can represent a word using word-embedding methods, such as Word2Vec (Mikolov et al., 2013) or GloVe (Pennington et al., 2014), a single character using one-hot encoding or character-embedding (Char2Vec), encoded features, or contextual embeddings (e.g., BERT (Devlin et al., 2018), RoBERTa (Liu, Ott, Goyal, Du, Joshi, Chen, & Stoyanov, 2019), XLNet (Yang, Dai, Yang, Carbonell, Salakhutdinov, & Le, 2019)) based on the surrounding words.
- o Number of layers and the DNN information flow – for instance, encoder-decoder architecture (Cho et al., 2014).
- o Activation functions (the neuron's function) – including Sigmoid, Tan-h, ReLU, and Softmax.
- o Loss functions (the "size" of the training error) – regression tasks can be: i) mean squared error (MSE), ii) mean squared logarithmic error, iii) mean absolute error; for binary classification tasks: i) binary cross-entropy, ii) hinge, iii) squared hinge; for multi-class classification: i) multi-class cross-entropy, ii) sparse multi-class cross-entropy, iii) Kullback- Leibler divergence.

- Hyper-parameters:
  - o Type and size of the regulation layers – regulation layers reduce overfitting by adding constraints to the DNN. These constraints, such as dropout (Srivastava et al., 2014), L1, and L2, prevent the model from learning the training data and force it to learn the patterns in the data.
  - o Batch size - the number of examples to use in a single training pass.
  - o Number of epochs and the epochs' size - the number of iterations on the training data and the number of examples to use during the entire training process.
  - o Learning rate, method, and configuration - such as stochastic gradient descent (SGD), adaptive moment estimation (Adam) (Kingma & Ba, 2014), and Adagrad (Duchi, Hazan & Singer, 2011).

Theoretically, architecture components are also hyper-parameters. However, from a practical perspective, once architecture components are chosen, they are usually fixed. There are techniques that can be applied to find and set these architecture components and hyper-parameters automatically. These techniques are called AutoML and are suitable for many different DNN models (and classical

ML models). However, AutoML has its limitations: it is often costly (training the model repeatedly), does not fit large-scale problems, and may lead to overfitting (Feurer & Hutter, 2019). It is advisable to check AutoML optimization methods such as submodular optimization (Jin, Yan, Fu, Jiang, & Zhang, 2016), grid search (Montgomery, 2017), Bayesian optimization (Melis, Dyer, & Blunsom, 2017), neural architecture search (So, Liang, & Le, 2019), and others (Feurer et al., 2019) or, if the researcher has a hypothesis or intuition about the problem, it is also possible to test multiple architecture components and hyper-parameters combinations manually. Moreover, training a large DNN language model such as a BERT-based model with standard pre-defined hyper-parameters on public cloud servers costs $2,074-$12,571, depending on the hyper-parameters and the corpus size (Devlin et al., 2018), while using neural architecture search (So et al., 2019) to train a DNN language model with hyper-parameters optimized for the specified task costs $44,055–$3,201,722 (Strubell, Ganesh, & McCallum, 2019). Therefore, the budget is another consideration for using some AutoML methods.

Numerous DH studies have demonstrated the importance and the impact of hyper-parameters optimization on the DNN accuracy. Tanasescu, Kesarwani, & Inkpen (2018) optimized hyper-parameters for poetic metaphor classification. They experimented with different activation functions (ReLU, Tan-h for the inner layers and Softmax and Sigmoid for the output layer), number of layers (1-4), number of neurons in each layer (6-306), dropout rate(0-0.9), number of epochs (20-1000), and batch size (20-200). The optimization increased the metaphor classification F-score by 2.9 (from 80.4 to 83.3) and precision by 5.6 (from 69.8 to 75.4). Wang et al. (2016), used a DNN model for Chinese song iambics generation and tested several architecture components. In their research, Wang et al. (2016) added an attention layer (Bahdanau et al., 2015) on top of bidirectional LSTM layers and tested several domain-specific training methods. This DNN domain optimization made it possible to achieve near-human performance. These use-cases emphasize how important it is for DH/LIS experts to understand architecture components and hyper-parameters and their usage.

**Domain-specific Dataset Adaptation for DNN**

Using DNN models in some domains can also require adaptation of the data (preprocessing) prior to inputting it into the DNN model. A use-case study of Won, Murrieta-Flores, & Martins (2018) aimed to perform Named Entity Recognition (NER) on two historical corpora, Mary Hamilton Papers (modern English from 1750 to 1820) and the Samuel Hartlib collection (early modern English from 1600 to 1660). NER is an NLP task which outputs identification of entity types in text. Entity types can be places, people, or organization names and other "known names". The historical corpus selected in Won et al. (2018) was OCRed and preserved in hierarchical XML files with texts and metadata. DNN models (and the tools used in the study) for NER are not designed to work directly on XML since XML is a graph-based format, and NER is a sequence-based task. It should be noted that there are graph-based DNN models (e.g., Scarselli, Gori, Tsoi, Hagenbuchner, & Monfardini, 2008), but they are not suitable for the NER task. Therefore, Won et al. (2018) needed to adapt their domain data by "translating" the XML markup into text sequences that a DNN model can receive as input. In this preprocessing phase, the researchers took into account the metadata that exists in the domain that was embedded in the XML file, such as authorship, dates, information about the transliteration project, corrections and suggestions made by the transliterators, and particular words and phrases annotated within the body text. Moreover, the square brackets (and their content) added by the transcribers were semi-automatically removed from the text. The metadata was added to the text sequence as labels for the training data to improve the accuracy of the results. Won et al. (2018) did not use DNN models directly but rather used "off the shelf" software to conduct their research. However, they concluded the research with the recognition that using pre-made tools is not sufficient - "*Finally, it must be noted that although this research accomplished the evaluation of the performance of these NER tools, further research is needed to deeply understand how the underlying models work with historical corpora and how they differ.*"

**DNN Pipeline Adaptation**

DNN models are designed to work in a certain pipeline of components to solve a specific task. For example, a "naïve" DNN based pipeline for the OCR of a book collection will be: 1) scan a book page, 2) use the image as an input to an image-to-text DNN model, 3) use the obtained text or post-process it to correct errors. However, in some cases, it is advisable to design a new domain-specific pipeline to solve the task or increase the model's accuracy. A use-case of such a domain-specific OCR pipeline is presented by Cilia et al. (2020). The goal of the study was identification of the page's writer for each page of the given medieval manuscript. Medieval handwritten manuscripts present two unique challenges for OCR: 1) first section letters or titles may be drawn as a picture over several lines, and 2) handwritten lines are not always aligned and may reduce accuracy when performing a full-page OCR. Cilia et al. (2020) designed a pipeline for processing handwritten medieval texts with three main steps, using: 1) an object detector to detect lines in the page's scanned image and separate a picture at the top from the text lines, 2) a separate DNN classifier to classify each line, and 3) a majority vote among multiple DNN classifiers obtained for each line and picture object at the line-level, in order to make a decision for the classification (writer identification) of the entire page. This pipeline, tailored to the medieval paleography domain, solved the domain's unique challenges by separating between picture objects and text lines and classifying each line with a different classifier instead of classifying an entire page with a single DNN model (the naïve pipeline). This pipeline's domain adaptation approach combined with the transfer learning approach, described in the previous section, produced an impressive 96% accuracy in identifying writers that would not have been achieved without this adaptation.

Pipeline adaptation is not just pipelining different models or combining ML and DL; it is also re-training and adapting an existing model, i.e., fine-tuning a model. Fine-tuning a model is a subset of transfer learning, in which a model is trained on a different dataset and also changed by setting different hyper-parameters or adding new last layers on top of the model to fit a specific task. In their research, Todorov and Colavizza (2020), fine-tuned a BERT-based model (Devlin et al., 2018) for increasing the annotation accuracy of NER in French and German historical corpora. In particular, the Groningen Meaning Bank's Corpus Annotated for NER was applied (Bos, Basile, Evang, Venhuizen, & Bjerva, 2017). To embed words (including sub-words) and characters, four models were applied: (1) newly trained word-embeddings on their historical corpus, (2) in-domain pre-trained embeddings that were trained on another corpus in the same domain, (3) BERT-based embedding that was trained on French and German Wikipedia, and (4) character level embeddings learned from the historical corpus training data. As can be observed from Figure 2, Todorov et al. (2020) combined the embedding (by concatenation) and transferred the unified embeddings to a new layer based on a Bi-LSTM-CRF layer. A Bi-LSTM-CRF layer is a Bidirectional (Schuster et al., 1997) Long Short-Term Memory (Hochreiter et al., 1997) layer that merges the sub-word embedding input into a word-level output and transfers its output to fully connected layers (one layer per each entity type) which then outputs tag (entity type) probabilities for each token using Conditional Random Fields (Lafferty, McCallum, & Pereira, 2001). The Bi-LSTM-CRF method has been shown as useful and accurate by Lample, Ballesteros, Subramanian, Kawakami, & Dyer (2016). They also changed the LSTM activation function (remove the tan-h function) and tried three different hyper-parameters configurations. Using the domain-specific pipeline, model, and hyper-parameters, the researchers dramatically increase the accuracy (in some entity types by over 20%) of NER task on French and German historical corpora compared to a state-of-the-art baseline model. Moreover, they tested the impact of the pre-trained generic embedding. They found that (1) without using the open-domain embedding (BERT), their model did not attain high accuracy, and (2) on the other hand, "freezing" the open-domain embedding layers (i.e., using them but re-training only the top layers on the domain-specific historical data) did not affect the accuracy. These findings demonstrate the importance of adapting DNN models to a specific domain and task, while reducing the training time and costs by freezing the large open-domain layers. It is essential to note that besides inputting the historical corpora documents into the DNN model, Todorov et al. (2020) also tested the addition of manually-created features to the documents such as title, numeric and other markups; these features did not have any effect on the accuracy, proving that the DNN model "learned" (or at least did not need) these features.

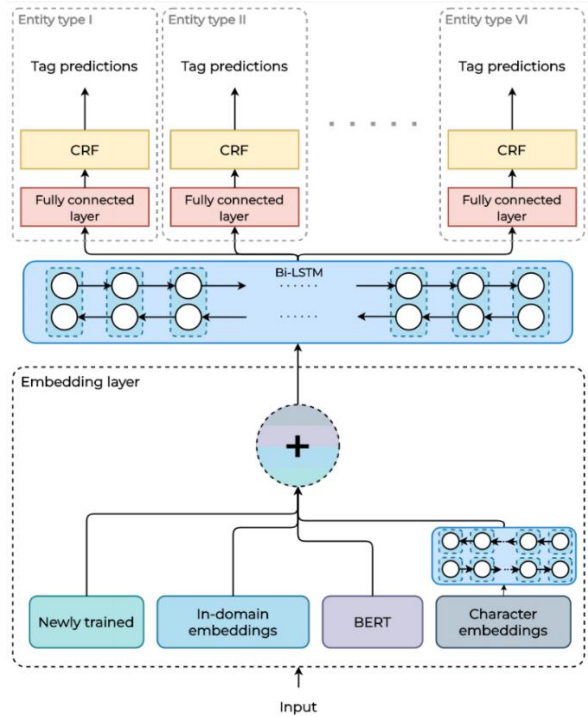

*Figure 2: Historical corpora NER fine-tuning pipeline (Todorov & Colavizza, 2020).*

## A Decision Model for Using Deep Learning for Digital Humanities Research

Based on the above analysis of challenges and possible solutions illustrated by multiple use-case studies described in the recent literature, it is clear that the DH/LIS experts must know just enough math, understand the inner-working of ML and DL algorithms, Python programming, and use these frameworks and other popular modules (Géron, 2019).

Therefore, this paper argues that DH/LIS researchers can no longer see NLP and ML researchers as their "tool makers", and must learn to apply and adapt deep learning models (DNNs) to their specific research domain. However, since working with DNN models requires significant effort, computational resources, budget, and time, a decision model was formulated for assisting DH experts in determining when it is "worthwhile" to invest in training DNN models. The decision model is based on two strategies: 1) the data availability strategy – how to assess the types of methods and models suitable for the available dataset, and 2) the domain adaptation strategy – how to determine whether and when it is "worthwhile" to invest in domain adaptation.

Figure 3 presents the data availability strategy and leads to three possible recommendations: (1) with no data, either zero-shot DL models, or hard-coded rules/assumptions regarding domain data should be implemented, based on prior knowledge and experience; (2) with limited data, either classical machine learning algorithms, such as SVM or HMM, or few-shot DL models can be used; otherwise (3) it is advisable to use supervised deep learning models for the task. It should be noted that if the

DNN model is overfitting (high accuracy on the training dataset and low accuracy on the validation dataset), it is advisable to increase the dataset size by employing expert workers, crowdsourcing, or synthetic data generation. Figure 4 presents the domain adaptation strategy and also leads to three possible recommendations: (1) if strict rules can be defined, there is no need for ML or DL; (2) with limited resources or for low accuracy tasks, ML is the preferable option, and (3) with the appropriate resources and a need for high accuracy, DL with domain adaptation should be utilized. A researcher can use both strategies of the proposed decision model to choose the recommended approach for the given task. Since there are many different text analysis tasks, some aspects of the strategies depend on the expert's assessment; for example, "what is considered a small or a large dataset?" and "what is low or high accuracy?". These assessments should be performed by the researcher based on the concrete task, domain, and needs. Notice that the advice to use DNN models does not mean that it is not recommended to combine them with ML algorithms when suitable.

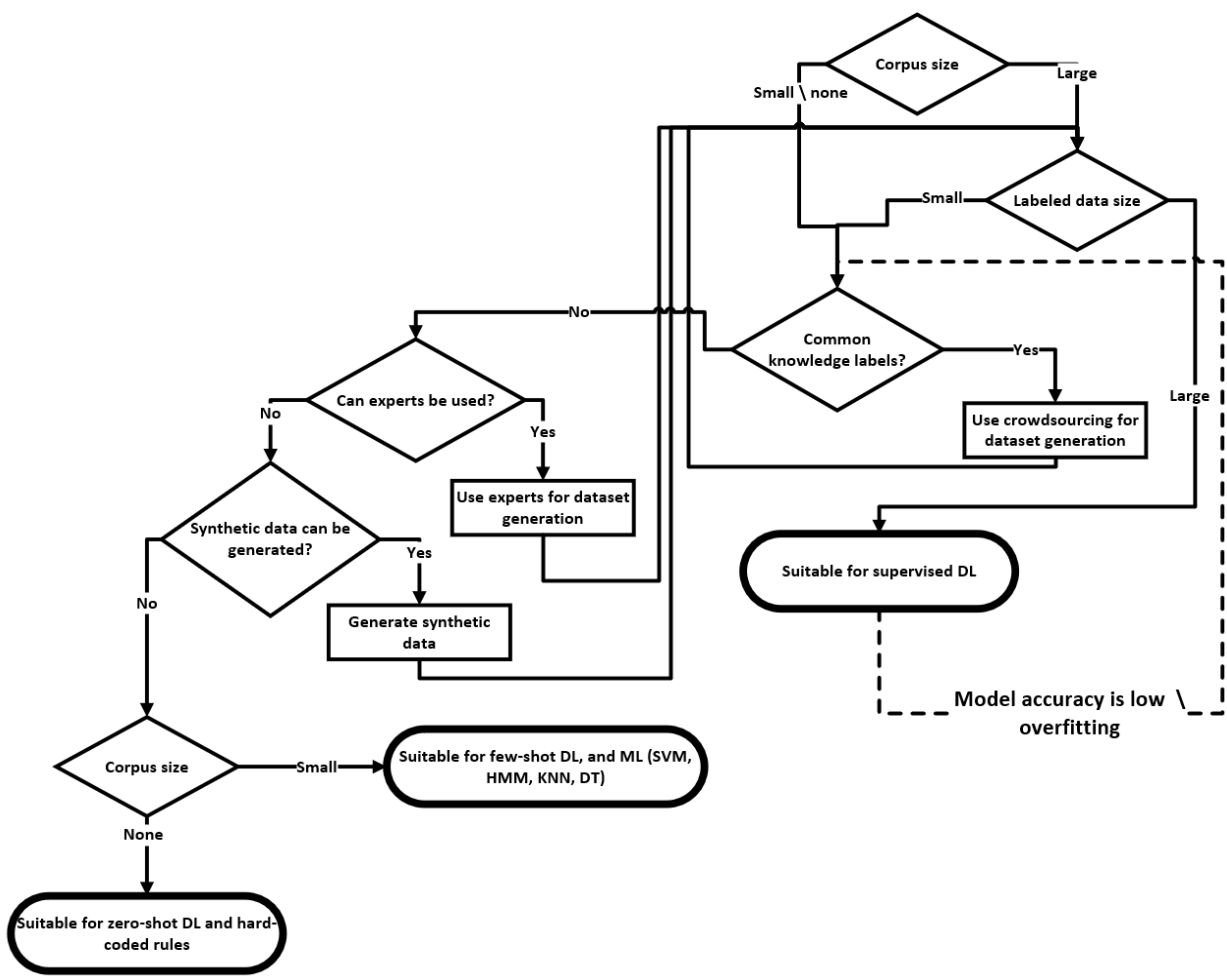

*Figure 3: Data availability strategy for DH researchers*

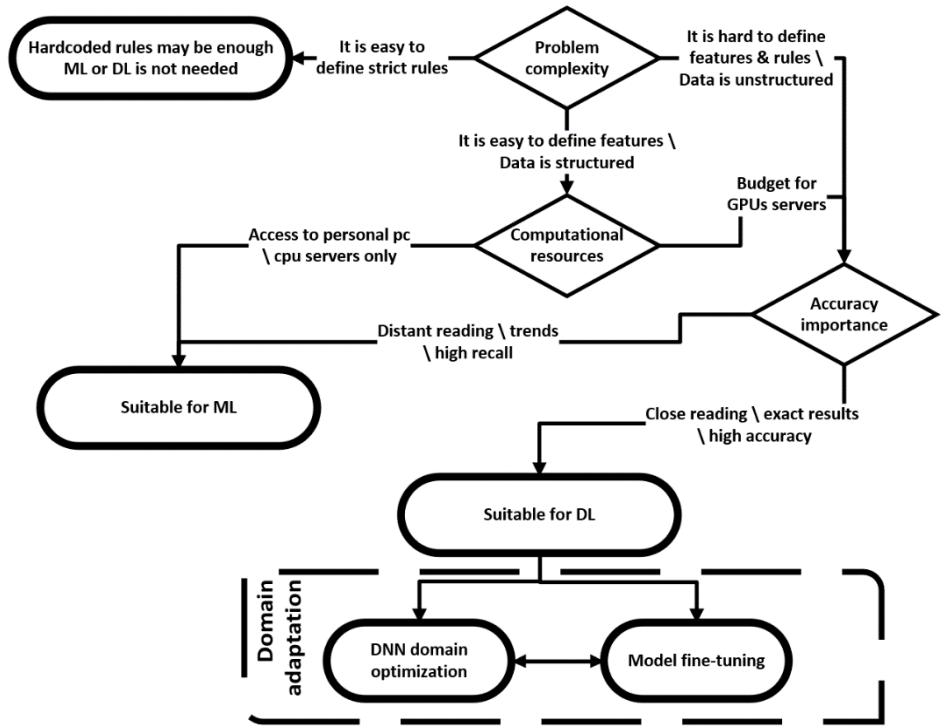

*Figure 4: Domain adaptation strategy for DH researchers*

As can be observed from the proposed decision model, supervised DL should be used when there is a large corpus (or a large corpus can be generated), for complex problems such as unstructured texts, when the researcher has a budget for computational resources (GPUs servers), and accuracy is essential (domain adaption is always assumed). Since most of the DH corpora are not labeled, dataset generation will most probably be required. When the labeling requires only "common knowledge", it is advisable to use crowdsourcing (if possible); otherwise, the researcher should consider using domain experts or automatic generating of synthetic data as explained above in this paper. A step-by-step example for decision model usage for a specific DH task can be found in Appendix II.

It should be noted that the extensive computational resources needed to train DNN models have an impact on the environment. DL may become a major contributor to climate change if the exponential growth of training more and more DNN models continues (Anthony, Kanding, & Selvan, 2020; Hsueh, 2020). It has been estimated that training one transformer model such as BERT-based (Devlin et al., 2018) will produce similar amounts of $CO_2$ to those of air travel of one person from NY to SF; using a neural architecture search (So et al., 2019), an AutoML method, will produce almost five times more $CO_2$ than an average car produces throughout its lifetime including the fuel (Strubell et al., 2019). We note that the proposed decision model does not consider environmental impact, yet researchers should be aware of this and take it into consideration.

By using this decision model as a guideline and applying the suggested solutions for the two fundamental challenges faced by many DH projects – DH-specific training dataset generation and model adaptation, DH/LIS experts can solve a variety of important tasks in the field for diverse national languages, such as 1) improving OCR post-correction (including restoring damaged text); 2) automated ontology and knowledge graph construction for various DH domains (based on entity/category and relation extraction and NER); and 3) corpus-based stylometric analysis and profiling of DH resources (e.g., identification of an author, date, location, and sentiment of the given text or image).

## Conclusion and Discussion

This paper presents the main two challenges almost every DH/LIS research can expect to encounter using DNN models in her research. Although classic learning techniques based on rules, patterns, or predefined features are no longer considered state-of-the-art in many text processing tasks (e.g., Thyaharajan, Sampath, Durairaj, & Krishnamoorthy, 2020; Glazkova, 2020), DH/LIS researchers are still using them often, even when there is a better alternative such as deep neural networks. The reasons for avoiding using deep learning in DH may be the lack of "off-the-shelf" tools tailored for the specified task, lack of training data, as well as time, computational resources, and budget limitations. Based on the presented investigation of the main challenges of using DNN in DH research and the proposed decision model for handling these challenges, and the potential adoption of DNN methods, this paper argues that DH/LIS researchers should expand their arsenal of computational skills and methods. A DH expert must acquire in-depth knowledge in mathematics, software programming and have a deep understanding of the usage of deep neural network frameworks. Therefore, we encourage DH/LIS academic departments to introduce the following topics into their academic syllabus, at the applied (rather than theoretical) level:

- Multivariable calculus (partial derivatives, gradients, high order derivatives),
- Linear algebra (vector space, matrices operations, matrices decompositions),
- Probability (distribution, entropy),
- Statistics (bayesian, parameter estimation, overfitting, and underfitting),
- Mathematical optimization (gradient descent, stochastic gradient descent),
- Unsupervised machine learning (k-means, hierarchical clustering, local outlier factor),
- Supervised machine learning (SVM, logistic regression, naïve bayes, knn),
- Unsupervised and self-supervised deep learning (autoencoders, deep belief networks, generative adversarial networks, embeddings),
- Supervised deep learning (feed-forward, RNN, Self-Attention Network (SAN), CNN,

- Python / R programming (working with data, visualization, ML and DL frameworks, working with GPUs).

Adding these topics to the academic syllabus of DH/LIS experts does not mean that DH/LIS experts will become Computer Science experts, but rather they will be able to comprehend and adapt DL algorithms for their needs. Using this knowledge, DH/LIS experts will no longer be limited to "off the shelf" tools developed for generic open-domain tasks, and will be able to utilize the full potential of the DL algorithms.

Finally, in addition to raising awareness of digital humanities researchers of deep neural networks as the state-of-the-art text analysis method, researchers should be encouraged to generate and release public DH/LIS corpora for training deep neural networks.

# Appendix I - Technical Description of DNN Models for Text Analysis

A deep neural network is a computational mathematical model that consists of mathematical functions (neurons) arranged in layers. Each neuron is a function, mapping a certain set of input values to output values, with the neuron forwarding the computed result to the neurons in the next layer. Each neuron has a "weight" value representing the neuron's influence on the output. If a neuron's weight is 0, it does not affect the output, and the larger its weight, the larger its influence. The actual learning of the DNN model is learning the "right" weights for the task. The model accepts input in the first layer and passes it from layer to layer until it becomes the output in the network's last layer. In the first learning cycle (batch), the neurons' weights are initialized with semi-random values. At the end of each learning cycle, the gap between the output values and the expected values ("loss") is calculated. Using the loss value, the neurons' weights are updated by applying the backpropagation method (Rumelhart, Hinton, & Williams, 1986). When the loss is small, the network learns "right", and as the loss increases, the network learns "wrong". A dense layer is a layer where all its nodes are connected to the next layer's nodes. Many neural network parameters and architecture components, such as the number of layers, number of neurons in a given layer, batch size, learning rate, and others, cannot be learned by the network itself and must be predefined and tuned to the specific task or domain. To tune or optimize the hyper-parameters, the network should be repeatedly trained with some initial hyper-parameters values and evaluated with a validation set.

Since DNN models are a set of chained mathematical formulas, the textual resources should be encoded into numeric values before input into the DNN model. The inputs can be encoded into vectors (a set of numbers) using embedding methods such as Word2Vec (Mikolov, Sutskever, Chen, Corrado, & Dean, 2013) or GloVe (Pennington, Socher, & Manning, 2014). A Word2Vec is actually a DNN model that learns how to represent words in meaningful vectors in order to capture the semantic and syntactic similarities and relations between words. Each word is encoded using one-hot-encoding, and the network learns one of two tasks: (1) Common Bag Of Words – the network receives a word and tries to predict the next word in the sentence or (2) Skip-Gram – the network receives a word and tries to predict the probability of each word in the corpus being the next word. When trained on large corpora, the Word2Vec hidden (single) layer actually represents the language model of that corpora and has proven to be a good method for encoding text into vectors (Mikolov et al., 2013).

Over the years, different deep learning layers have been developed for various NLP tasks. Common Recurrent Neural Network (RNN) layers such as Long Short-Term Memory (LSTM) (Hochreiter & Schmidhuber, 1997) and Gated Recurrent Units (GRU) (Cho, van Merrienboer, Gulcehre, Bahdanau, Bougares, Schwenk, & Bengio, 2014) allow the layer to "remember" previously encountered data and thus learn based on a broader view of the data (for example, learning from the entire sentence and not just from a specific word). Bidirectional-RNN (Schuster & Paliwal, 1997) is a technique allowing layers such as LSTM and GRU to use the input data in the direct and reverse learning order based on the following data (for example, words) rather than only by the preceding data. These layers are used to construct different architectures, including a sequence-to-sequence model (Sutskever, Vinyals, & Le, 2014) that uses an encoder-decoder (Cho et al., 2014) architecture. This model maps a sequence input (e.g., a sentence) to a sequence output. A dropout regulation layer is usually added (Srivastava, Hinton, Krizhevsky, Sutskever, & Salakhutdinov, 2014) to prevent overfitting (a situation where the network learns the training examples but is unable to predict with the same accuracy for new examples) and overcome the vanishing gradient problem (Hochreiter, 1998). In each learning cycle, several neurons are "dropped out" (setting their weight to 0) so that the network must learn to generalize its predictions instead of learning a simple mapping between the input and the output values. However, the RNN encoder-decoder architecture suggested by Cho et al. (2014) is less practical for long texts (i.e., sequences), since it uses a fixed-length context vector (i.e., the encoder output vector for the decoder). To overcome this limitation, Bahdanau et al. (2015) suggested an attention model that assigns a weight for each token (i.e., word) in the sequence (i.e., sentence), allowing the model to focus on important tokens instead of encoding the whole sequence. A Self-Attention Network (SAN) (Vaswani et al., 2017) is a model where the attention is not a separate layer on top of the other layers in the model, but is embedded into the model's layers (e.g., instead of the attention to "look" at the output of the previous layer, it "looks" at the inputs of the current layer), and thus can be added several times inside the model.

The current state-of-the-art approach for many NLP tasks is the Transformer architecture (Vaswani et al., 2017). The Transformer architecture is based on several dense layers (instead of RNNs) with self-attention. The well-known Bidirectional Encoder Representations from Transformers (BERT) (Devlin, Chang, Lee, & Toutanova, 2018) and the GPT-2 (Radford et al., 2019) language models are based on the Transformer architecture. The main benefit of the transformers architecture is that it allows learning a contextual embedding of words. Unlike Word2Vec, each word may have a different vector depending on the words surrounding it. The BERT model achieves this contextual embedding using a specific training process. Instead of working at the word level (like Word2Vec), its training process includes two tasks at the sentence level: masked language modeling (MLM) and next sentence prediction (NSP). In the MLM task, for each input sentence, one or more words are masked, and the model's task is to generate the most likely substitution for each masked word. This is similar to "fill-in-the-blank" that children are asked to perform in elementary school. The model is forced to use the context words surrounding a mask to achieve high accuracy. In the NSP task, the model receives a pair of sentences and predicts if the second sentence follows the first one in the dataset. The combination of MLM and NSP training tasks allows modeling languages with an understanding of both word and sentence relations. Using a pre-trained BERT model trained on MLM and NSP tasks, researchers can add DNN layers on top of the model and fit it into a specific task (Devlin et al., 2018).

If the pre-trained model is large enough, Few-shot and Zero-shot methods can be used to reduce the need for large domain-specific training datasets for model's fine-tuning (Radford et al., 2019; Brown et al., 2020).

The vast amount of different training approaches, hyper-parameters tuning, and architectures illustrate the complexity and sensitivity of the models to a specific domain and sub-task.

Working with DNNs requires the researcher to be well acquainted with programming languages, such as Python (Rossum, 1995), and can be implemented using one of the common DNN frameworks: Tensorflow (Abadi, Barham, Chen, Chen, Davis, Dean, & Kudlur, 2016) with a popular abstraction module Keras (Gulli & Pal, 2017), PyTorch (Paszke, Gross, Massa, Lerer, Bradbury, Chanan, & Desmaison, 2019) or Theano (Bergstra, Breuleux, Bastien, Lamblin, Pascanu, Desjardins, & Bengio, 2010).

# Appendix II – A Step-by-step Example for Decision Model Usage

Consider the following case - a researcher aims to understand burial culture behavior and patterns in selected communities in Germany from the 18th century. However, she does not have annotated data for the task, but only a large corpus of tombstone images (e.g., Kollatz (2019)). The first step towards analyzing the burial culture in the selected communities is to create a structured dataset comprised of the information (person's names, family relationships, locations, dates, events) extracted from texts written on the tombstones. First, the researcher tries to use generic OCR software. However, it leads to low accuracy due to the uncommon fonts, a bias towards modern language, blurred and partially erased words in the images, and bias towards "white paper with text" (i.e., not suitable to find the tombstone in the image and extract the text from it and only from it). Then, following the domain adaptation strategy (Figure 4), it is clear that this problem (i.e., extracting the data from the tombstones) can not be solved using hardcoded rules, and it is even difficult to define features for classic ML algorithms for this task; therefore it seems that this research is suitable for DL. Figure 5 shows the chosen decision path on the diagram for the domain adaptation strategy (the blue line).

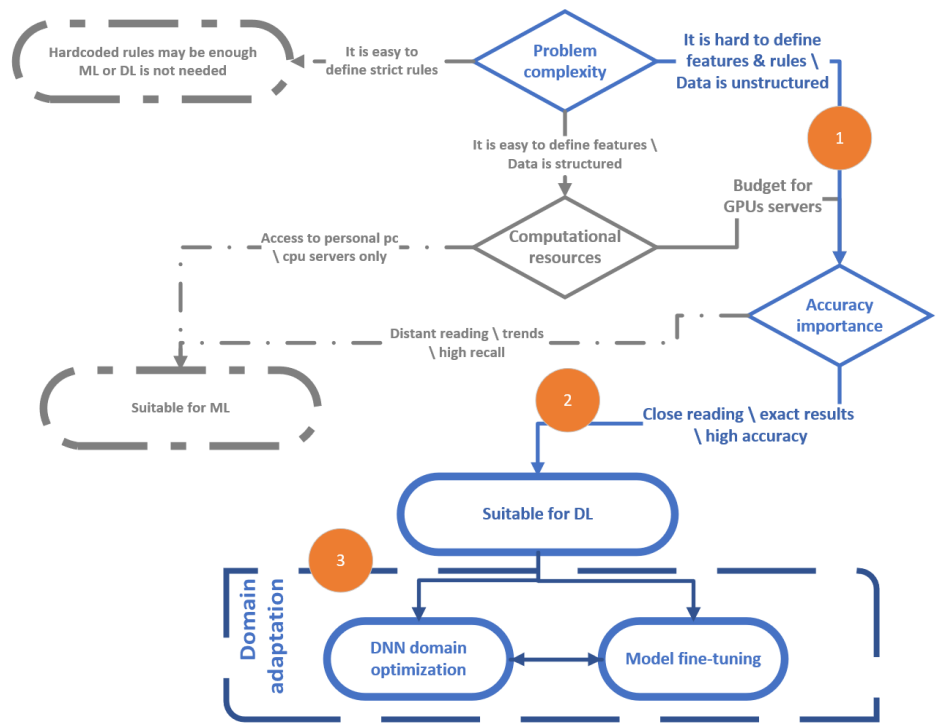

*Figure 5: The domain adaptation strategy chosen by the researcher (the blue lines)*

Before starting the domain adaptation process, the researcher follows the data availability strategy (Figure 3). Since the researcher has no training data at all (no annotated images of tombstones or corresponding metadata), based on the data availability decision strategy, she first tries to use crowdsourcing to annotate the data on a subset of a few hundred tombstones images. Using the annotated dataset, the researcher fine-tunes and optimizes a DNN model to extract the data from the images. However, the model's accuracy is not satisfactory. After examining the DNN's wrong outputs, the researcher decides that the crowdsourcing quality is the problem. Therefore, the researcher asks three experts in the field to annotate a hundred additional tombstone images. Then she fine-tunes the DNN model again and examines the obtained accuracy. It seems that the model had achieved high accuracy on the training dataset; however, on the validation dataset (i.e., examples from the dataset not used during training), the model fails to achieve similar accuracy (i.e., overfitting). The researcher hypothesized that her model was unable to achieve high accuracy on the validation dataset since it was not successful in overcoming the bias towards modern language (i.e., to identify entities that are not used in modern languages) and the "white paper with text" bias (i.e., to find the tombstone in the image and extract the text only from it). The researcher decides to generate a synthetic dataset using corpora annotated for Named Entity Recognition (NER) (e.g., Neudecker, 2016) from the same period and location (i.e., 18th-century German books) and trains a new DNN model for NER of typical entities (e.g., names, locations, events, and date formats). Using the NER DNN model, the researcher generates a list of fictitious entities. With this list, the researcher generates tens of thousands of tombstone images by placing the entities on tombstone images using graphic editing software. Figure 6 presents the process the researcher has executed using the data availability strategy.

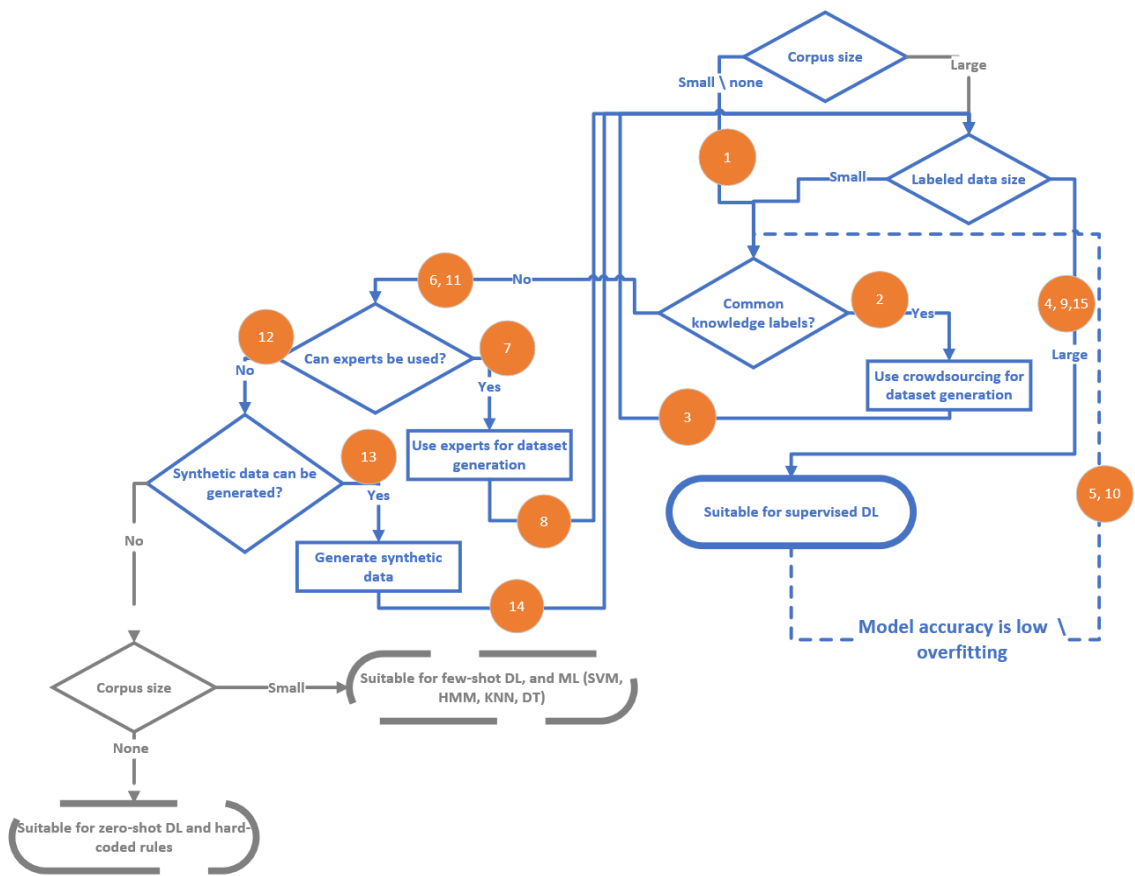

*Figure 6: The data availability strategy chosen by the researcher (the blue lines)*

Then, using the new large synthetic dataset of tombstones images, the researcher trains the DNN model again, and this time the results are much better. However, the researcher still wishes to improve the accuracy by tweaking hyperparameters, such as batch size and dropout rate. Finally, the model's accuracy is satisfactory, and the researcher can investigate the community burial culture.

# Appendix III

The DH journals:

1. Frontiers in Digital Humanities
2. Digital Scholarship in the Humanities
3. Digital Humanities Quarterly
4. Digital Studies
5. Journal of Digital Humanities
6. Journal of Cultural Analytics
7. Journal of Data Mining and Digital Humanities
8. International Journal of Digital Humanities
9. Journal on Computing and Cultural Heritage
10. Journal of the Text Encoding Initiative
11. International Journal of Humanities and Arts Computing
12. Digital Medievalist

The Bioinformatics journals:

1. Bioinformatics
2. PLOS Computational Biology
3. Briefings in Bioinformatics
4. BMC Bioinformatics
5. GigaScience
6. Journal of Theoretical Biology
7. Journal of Cheminformatics
8. IEEE/ACM Transactions on Computational Biology and Bioinformatics
9. Genomics, Proteomics & Bioinformatics
10. BMC Systems Biology
11. Mathematical Biosciences
12. Journal of Mathematical Biology